%% file: egpaper_final.tex
\documentclass[10pt,twocolumn,letterpaper]{article}

\usepackage{cvpr}
\usepackage{times}
\usepackage{amsmath}
\usepackage{amssymb}
\usepackage{color}
\usepackage{times}
\usepackage{graphicx}
\usepackage{array}
\usepackage{url}
\usepackage{booktabs}
\usepackage{xcolor}
\usepackage{cite}
\usepackage{multirow}
 \usepackage{comment} 

\usepackage{dsfont}

\usepackage{soul}

\newcommand{\htg}{heterogeneous\ }
\newcommand{\Htg}{Heterogeneous\ }
\DeclareMathOperator*{\argmax}{arg\,max}

\usepackage[breaklinks=true,bookmarks=false]{hyperref}

\cvprfinalcopy 

\def\cvprPaperID{1022} 

\begin{document}

\title{Heterogeneous Memory Enhanced Multimodal Attention Model for \\Video Question Answering}

\author{Chenyou Fan$^{1,*}$, Xiaofan Zhang$^1$, Shu Zhang$^1$, Wensheng Wang$^1$, Chi Zhang$^1$, 
Heng Huang$^{1,2,*}$
\\$^1$JD.COM, $^2$ JD Digits\\
{\tt\small $^{*}$chenyou.fan@jd.com, $^{*}$heng.huang@jd.com}}

\maketitle

\begin{abstract}
In this paper, we propose a novel end-to-end trainable Video Question Answering (VideoQA) framework with three major components: 1) a new heterogeneous memory which can effectively learn global context information from appearance and motion features; 2) a redesigned question memory which helps understand the complex semantics of question and highlights queried subjects; and 3) a new multimodal fusion layer which performs multi-step reasoning by attending to relevant visual and textual hints with self-updated attention. Our VideoQA model firstly generates the global context-aware visual and textual features respectively by interacting current inputs with memory contents. After that, it makes the attentional fusion of the multimodal visual and textual representations to infer the correct answer. Multiple cycles of reasoning can be made to iteratively refine attention weights of the multimodal data and improve the final representation of the QA pair. Experimental results demonstrate our approach achieves state-of-the-art performance on four VideoQA benchmark datasets.
\end{abstract}

\input{intro}

\input{related}

\input{approach}

\input{experiments}
\input{conclusion}

{\small
\bibliographystyle{ieee}
\bibliography{egpaper_final}
}

\end{document}

%% file: intro.tex
\section{Introduction}
Video Question Answering (VideoQA) is to learn a model that can infer the correct answer for a given question in human language related to the visual content of a video clip. VideoQA is a challenging computer vision task, as it requires to understand a complex textual question first, and then to figure out the answer that can best associate the semantics to the visual contents in an image sequence.

\begin{figure}
\begin{center}
\includegraphics[width=1.\columnwidth, trim=5.5cm 4.cm 5.5cm 5.2cm,clip]{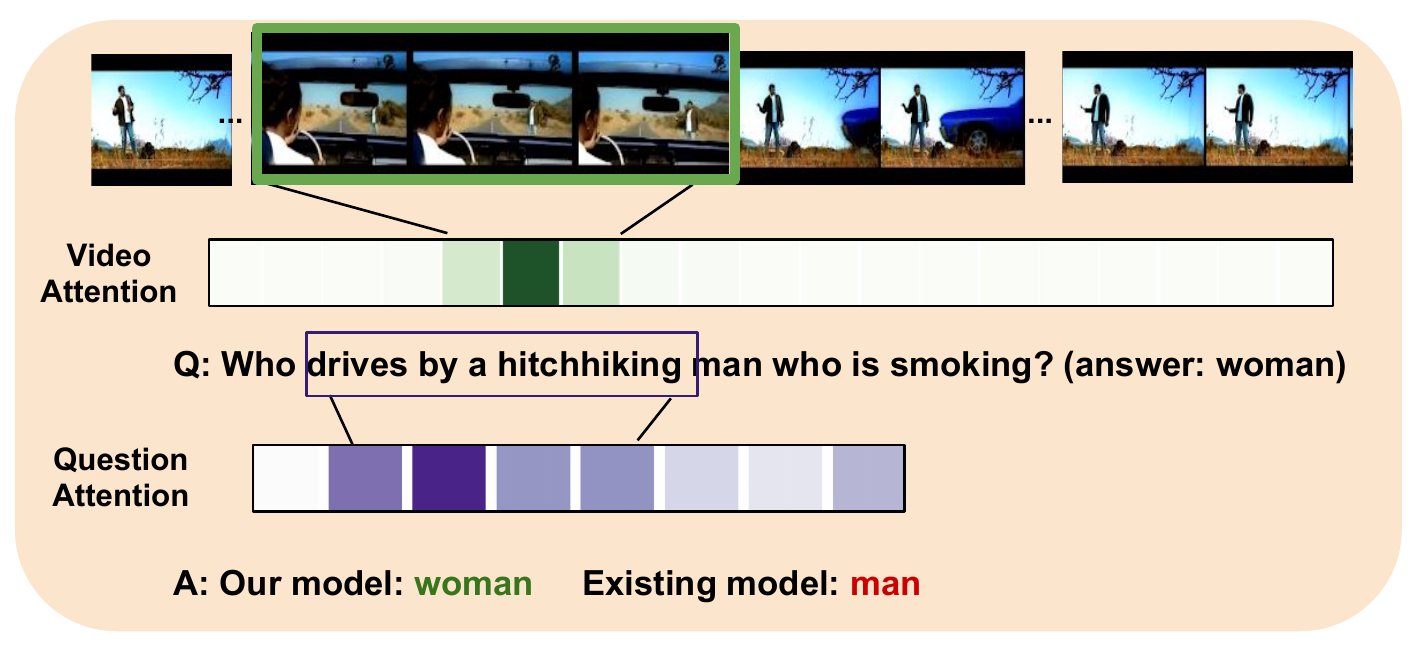}
\end{center}
\caption{VideoQA is a challenging task as it requires the model to associate relevant visual contents in frame sequence with the real subject queried in question sentence.
For a complex question such as ``Who drives by a hitchhiking man who is smoking?", the model needs to understand that the driver is the queried person and then localize the frames in which the driver is driving in the car.}
\vspace{-5pt}
\label{fig:demo}
\end{figure}

Recent work~\cite{yang2016stacked,andreas2016neural,anderson2017bottom,kembhavi2017you,ma2018visual,jang2017tgif} proposed to learn models of encoder-decoder structure to tackle the VideoQA problem. A common practice is to use LSTM-based encoders to encode CNN features of video frames and embeddings of question words into encoded visual sequence and word sequence. Proper reasoning is then performed to produce the correct answer, by associating the relevant visual contents with the question.
For example, learning soft weights of frames will help attend to events that are queried by the questions, while learning weights of regions in every single frame will help detect details and localize the subjects in the query. The former one aims to find relevant frame-level details by applying temporal attention to encoded image sequence~\cite{jang2017tgif,ma2018visual,xu2017video}. The latter one aims to find region-level details by spatial attention~\cite{xiong2016dynamic,kim2016multimodal,yang2016stacked,anderson2017bottom}.

Jang~\etal~\cite{jang2017tgif} applied spatiotemporal attention mechanism on both spatial and temporal dimension of video features. They also proposed to use both appearance (\emph{e.g.}, VGG~\cite{simonyan2014very}) and motion features (\emph{e.g.}, C3D~\cite{tran2015learning}) to better represent video frames. Their practice is to make early fusion of the two features and feed the concatenated feature to a video encoder. But such straightforward feature integration leads to suboptimal results.
Gao~\etal~\cite{gao2018motion} proposed to replace the early fusion with a more sophisticated co-memory attention mechanism.
They used one type of feature to attend to the other and fused the final representations of these two feature types at the final stage. However, this method doesn't synchronize the attentions detected by appearance and motion features, thus could generate  incorrect attentions. Meanwhile, this method will also miss the attention which can be inferred by the combined appearance and motion features, but not individual ones. The principal reason for the existing approaches to fail to identify the correct attention is that they separate feature integration and attention learning steps. \textbf{To address this challenging problem, we propose a new heterogeneous memory to integrate appearance and motion features and learn spatiotemporal attention simultaneously.} In our new memory model, the heterogeneous visual features as multi-input will co-learn the attention to improve the video understanding. 

On the other hand, VideoQA becomes very challenging if the question has complex semantics and requires multiple steps of reasoning. Several recent work~\cite{zeng2017leveraging,ma2018visual,gao2018motion} tried to augment VideoQA with differently embodied memory networks~\cite{weston2015mm,sukhbaatar2015end,xiong2016dynamic}.
Xu~\etal~\cite{xu2017video} proposed to refine the temporal attention over video features word by word with a conventional LSTM question encoder plus an additional LSTM based memory unit to store and update the attention. However, this model is easily trapped 
into irrelevant local semantics, and cannot understand the question based on the global context.
Both Zeng~\etal~\cite{zeng2017leveraging} and Gao~\etal~\cite{gao2018motion} used external memory (memory network~\cite{sukhbaatar2015end} and episodic memory~\cite{xiong2016dynamic} respectively) to make multiple iterations of inference by interacting the encoded question representation with video features conditioning on current memory contents. However, similar to many other work~\cite{xiong2016dynamic,jang2017tgif,anderson2017bottom}, the question representation used in these approaches is only a single feature vector encoded by an LSTM (or GRU) which lacks capability to capture complex semantics in questions such as shown in Fig.~\ref{fig:demo}. Thus, it is desired to design a new powerful model for understanding the complex semantics of questions in VideoQA. \textbf{To tackle this problem, we design novel network architecture to integrate both question encoder and question memory which can augment each other. The question encoder learns meaningful representation of question and the re-designed question memory understands the complex semantics and highlights queried subjects by storing and updating global contexts.}

Moreover, we design a multimodal fusion layer which can 
attend to visual and question hints simultaneously by aligning relevant visual contents with key question words.
After gradually refining the joint attention over video and question representations and fusing them with learned soft modality weights, the multi-step reasoning is achieved to infer the correct answer from the complex semantics.


Our major contributions can be summarized as follows: 1) we introduce a heterogeneous external memory module with attentional read and write operations such that the motion and appearance features are integrated to co-learn attention; 2) we utilize the interaction of visual and question features with memory contents to learn global context-aware representations; 3) we design a multimodal fusion layer which can effectively combine visual and question features with softly assigned attentional weights and also support multi-step reasoning; and 4) our proposed model outperforms the state-of-the-art methods on four VideoQA benchmark datasets.

%% file: related.tex
\begin{figure*}
\vspace{-15pt}
\begin{center}
\includegraphics[width=1.8\columnwidth, trim=1cm 3cm 0cm 0.8cm,clip]{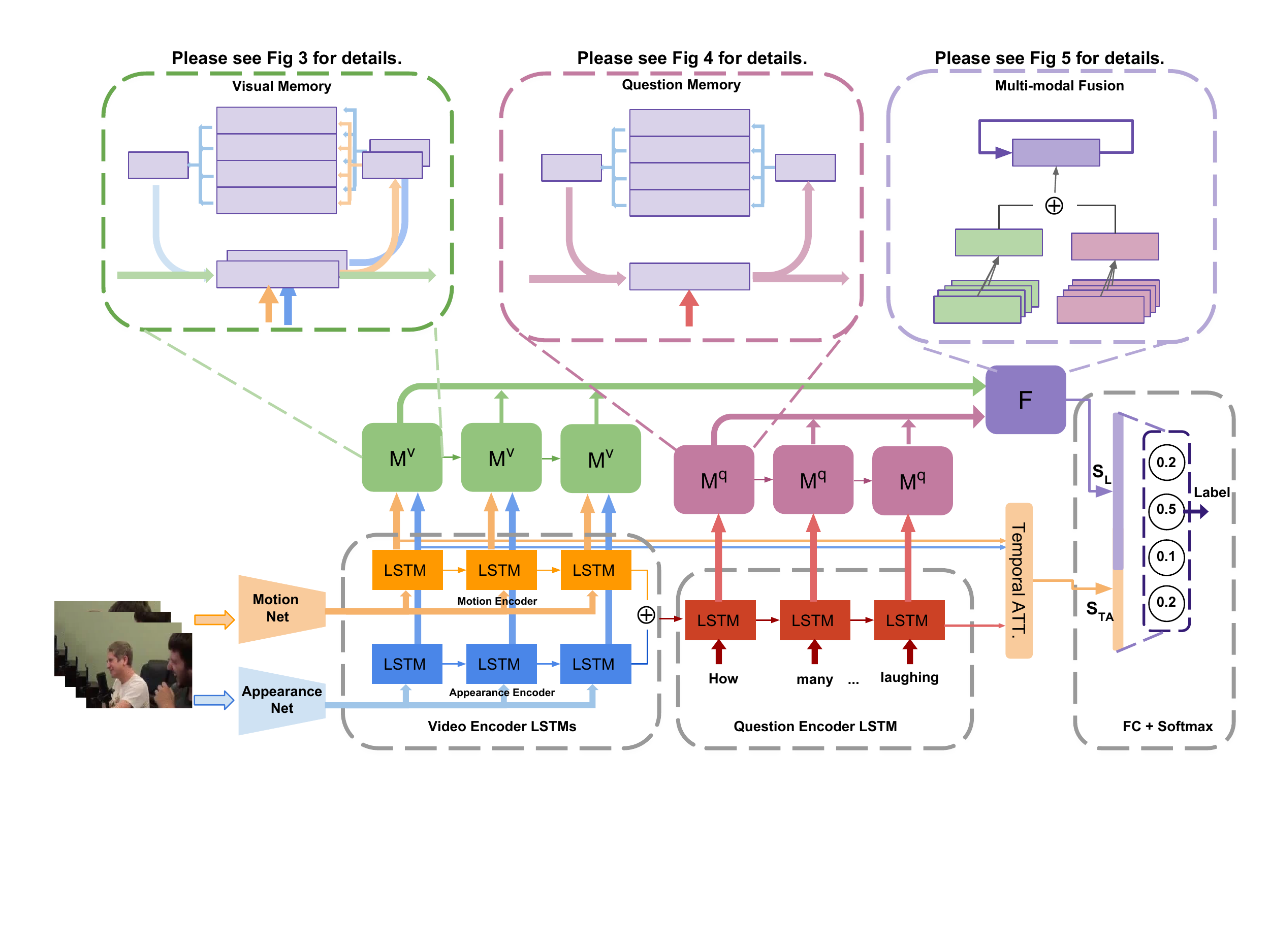}
\end{center}
\vspace{-15pt}
 \caption{Our proposed VideoQA pipeline with highlighted visual memory, question memory, and multimodal fusion layer.}
\label{fig:pipeline}
\vspace{-10pt}
\end{figure*}

\section{Related Work}
\textbf{Visual Question Answering (VQA)} is an emerging research area~\cite{malinowski2014multi,agrawal2015vqa,yang2016stacked,andreas2016neural,anderson2017bottom,kembhavi2017you,ma2018visual} to reason the correct answer of a given question which is related to the visual content of an image. 
Yang~\etal~\cite{yang2016stacked} proposed to encode question words into one feature vector which is used as query vector to attend to relevant image regions with stack attention mechanism. Their method supports multi-step reasoning by repeating the query process while refining the query vector. Anderson~\etal~\cite{anderson2017bottom} proposed to align questions with relevant object proposals in images generated by Faster R-CNN~\cite{ren2015faster} and compute the visual feature as a weighted average over all proposals.
Xiong~\etal~\cite{xiong2016dynamic} proposed to encode image and question features as facts and attend to relevant facts through attention mechanism to generate a contextual vector.  
Ma~\etal~\cite{ma2018visual} proposed a co-attention model which can attend to not only relevant image regions but also important question words simultaneously. They also suggested to use external memory~\cite{santoro2016meta} to memorize uncommon QA pairs.

\textbf{Video Question Answering (VideoQA)} extends VQA to video domain which aims to infer the correct answer given a relevant question of the visual content of a video clip. VideoQA is considered to be a challenging problem as reasoning on video clip usually requires memorizing contextual information in temporal scale. Many models have been proposed to tackle this problem~\cite{yu2016end,jang2017tgif,zeng2017leveraging,xu2017video,ye2017video,gao2018motion}. Many work~\cite{jang2017tgif,gao2018motion,ye2017video} utilized both motion (i.e. C3D~\cite{tran2015learning}) and appearance (i.e. VGG~\cite{simonyan2014very}, ResNet~\cite{he2016deep}) features to better represent video frames.
Similar to the spatial mechanism widely used in VQA methods to find relevant image regions, many VideoQA work~\cite{jang2017tgif,gao2018motion,ye2017video,xu2017video} applied temporal attention mechanism 
to attend to most relevant frames of a video clip.
Jang~\cite{jang2017tgif} utilized both appearance and motion features as video representations and applied spatial and temporal attention mechanism to attend to both relevant regions of a frame and frames of a video.
Xu~\etal~\cite{xu2017video} proposed to refine the temporal attention over frame features at each question encoding step word by word. 
Both Zeng~\etal~\cite{zeng2017leveraging} and Gao~\etal~\cite{gao2018motion} proposed to use external memory (Memory Network~\cite{sukhbaatar2015end} and Episodic Memory~\cite{xiong2016dynamic} respectively) to make multiple iterations of inference by interacting the encoded question feature with video features conditioning on current memory contents. 
Their memory designs maintain a single hidden state feature of current step and update it through time steps. However, this could hardly establish long-term global context as the hidden state feature is updated at every step. Neither are their models able to synchronize appearance and motion features.

Our model differs from existing work such that 1) we design a heterogeneous external memory module with attentional read and write operations that can efficiently combine motion and appearance features together; 2) we allow interaction of visual and question features with memory contents to construct global context-aware features; and 3) we design a multimodal fusion layer which can effectively combine visual and question features with softly assigned attentional weights and also support multi-step reasoning.

%% file: approach.tex
\section{Our Approach}
In this section, we illustrate our network architecture for VideoQA. We first introduce the LSTM encoders for video features and question embeddings. Then we elaborate on the design of question memory and \htg video memory. Finally, we demonstrate how our designed multimodal fusion layer can attend to relevant visual and textual hints and combine to form the final answer representation.

\subsection{Video and text representation} 
\textbf{Video representation.} Following previous work~\cite{jang2017tgif,xu2017video,gao2018motion}, we sample a fixed number of frames (\emph{e.g.}, 35 for TGIF-QA) for all videos in that dataset. We then apply pre-trained ResNet~\cite{he2016deep} or VGG~\cite{simonyan2014very} network on video frames to extract video appearance features, and use C3D~\cite{tran2015learning} network to extract motion features. We denote appearance features as $\mathbf{f}^a=[\mathbf{f}^a_1,\cdots,\mathbf{f}_{N_v}^a]$, and motion features as $\mathbf{f}^m=[\mathbf{f}^m_1,\cdots,\mathbf{f}_{N_v}^m]$, in which $N_v$ is number of frames.
The dimensions of ResNet, VGG and C3D features are 2048, 4096 and 4096.
We use two separate LSTM encoders to process motion and appearance features individually first, and late fuse them in the designed memory module which will be discussed in \S\ref{sec:htg}. In Fig.~\ref{fig:pipeline}, we highlight the appearance encoder in
blue and the motion encoder in orange. The inputs fed into the two encoders are raw 
CNN motion features $\mathbf{f}^m$ and appearance features $\mathbf{f}^a$, and the outputs are encoded motion and appearance features denoted as $\mathbf{o}^m=[\mathbf{o}^m_1,\cdots,\mathbf{o}_{N_v}^m]$ and $\mathbf{o}^a=[\mathbf{o}^a_1,\cdots,\mathbf{o}_{N_v}^a]$.

\textbf{Question representation.}
Each VideoQA dataset has a pre-defined vocabulary which is composed of the top $K$ most frequent words in the training set. The vocabulary size $K$ of each dataset is shown in Table~\ref{tab:data_stat}.
We represent each word as a fixed-length learnable word embedding and initialize with the pre-trained GloVe 300-D~\cite{pennington2014glove} feature. We denote the question embedding as a sequence of word embeddings $\mathbf{f}^q=[\mathbf{f}^q_1,\cdots,\mathbf{f}_{N_q}^q]$, in which $N_q$ is number of words in the question.
We use another LSTM encoder to process question embedding $\mathbf{f}^q$, as highlighted in red in Fig.~\ref{fig:pipeline}. The outputs are the encoded text features $\mathbf{o}^q=[\mathbf{o}^q_1,\cdots,\mathbf{o}_{N_q}^q]$ .

\begin{figure}
\vspace{-15pt}
\begin{center}
\includegraphics[width=0.9\columnwidth, trim=4.2cm 1.8cm 4.2cm 3.6cm,clip]{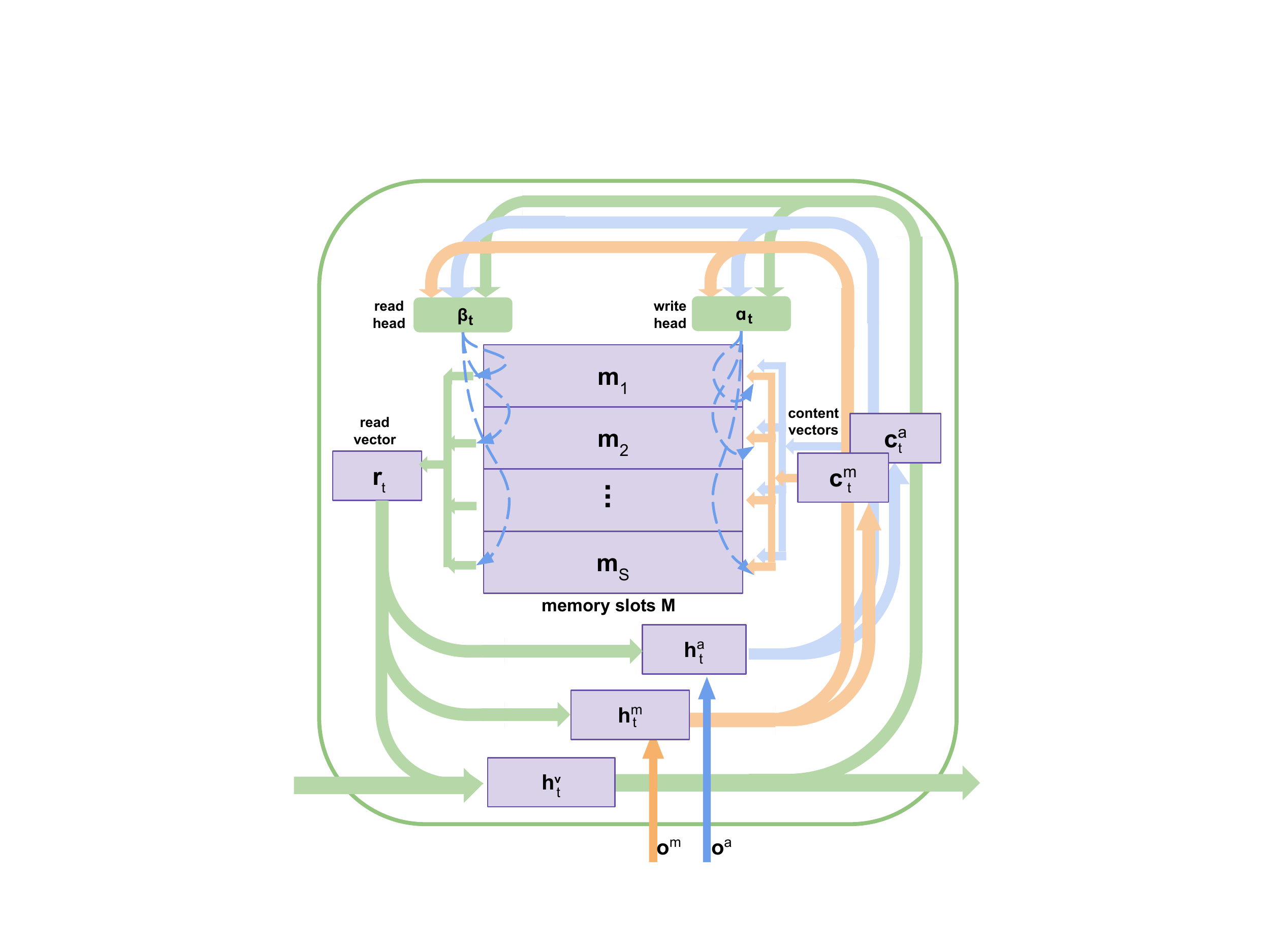}
\end{center}
 \caption{Our designed \htg visual memory which contains 
 memory slots $M$, read and write heads $\alpha,\beta$, and three hidden states $\mathbf{h}^m$, $\mathbf{h}^a$ and $\mathbf{h}^{v}$.
 }
 \vspace{-10pt}
\label{fig:memory-m2}
\end{figure}

\subsection{\Htg video memory}\label{sec:htg}

Both motion and appearance visual features are crucial for recognizing the objects and events associated with the questions. Because these two types of features are heterogeneous, the straightforward combination cannot effectively learn the video content. Thus, we propose a new heterogeneous memory to integrate motion and appearance visual features, learn the joint attention, and enhance the spatial-temporal inference. 

Different to the standard external memory, our new heterogeneous memory accepts multiple inputs including
encoded motion features $\mathbf{o}^m$ and appearance features $\mathbf{o}^a$, and uses multiple write heads to determine the content to write. 
Fig.~\ref{fig:memory-m2} illustrates the memory structure, which is composed of  memory slots $\mathbf{M}=[\mathbf{m}_1,\cdots,\mathbf{m}_S]$ and three hidden states $\mathbf{h}^m$, $\mathbf{h}^a$ and $\mathbf{h}^{v}$.
We use two hidden states $\mathbf{h}^m$ and $\mathbf{h}^a$ to determine motion and appearance contents which will be written into memory, and use a separate global hidden state $\mathbf{h}^{v}$ to store and output global context-aware feature which integrates motion and appearance information.
We denote the number of memory slots as $S$, and sigmoid function as $\sigma$. For simplicity, we combine superscript $m$ and $a$ for identical operations on both motion and appearance features.

\textbf{Write operation}. 
Firstly we define the motion and appearance content $\mathbf{c}_t^{m/a}$ to write to memory at $t$-th time as non-linear mappings from input and previous hidden state
\begin{equation} \label{eq:c_t}
\small
\mathbf{c}_t^{m/a} = \sigma(\mathbf{W}^{m/a}_{oc} \mathbf{o}_t^{m/a} + \mathbf{W}^{m/a}_{hc} \mathbf{h}_{t\text{-}1}^{m/a} + \mathbf{b}^{m/a}_c)
\end{equation}
Then we define $\small \boldsymbol \alpha_t^{m/a} \text{=}\{\alpha_{t,1}^{m/a},\ldots,\alpha_{t,S}^{m/a}\}$ as 
the write weights of $\mathbf{c}_t^{m/a}$ to each of $S$ memory slot given by
\begin{equation}  
\label{eq:softmax_a}
\small
\begin{split}
\mathbf{a}_t^{m/a} = &\mathbf{v}_a^\top \tanh( \mathbf{W}^{m/a}_{ca} \mathbf{c}_t^{m/a}  + \mathbf{W}^{m/a}_{ha} \mathbf{h}_{t\text{-}1}^{m/a} + \mathbf{b}_a^{m/a}) \\
&\alpha_{t,i}^{m/a} = \frac{\exp(a_{t,i}^{m/a})}{\sum_{j=1}^S \exp(a_{t,j}^{m/a})} \text{ \quad  for } i=1\ldots S 
\end{split}
\end{equation}
satisfying $\boldsymbol \alpha_t^{m/a}$ sum to $1$.
Uniquely, we also need to integrate motion and appearance information and make a unified write operation into current memory. Thus we estimate the weights $\boldsymbol \epsilon_t \in \mathbb{R}^3$ of motion content $\boldsymbol \alpha_t^{m}$, appearance content $\boldsymbol \alpha_t^{a}$ and current memory content $\mathbf{M}_{t\text{-}1}$ given by
\begin{equation} 
\small
\begin{split}
\mathbf{e}_t =&\mathbf{v}_e^\top \tanh(\mathbf{W}_{he} \mathbf{h}_{t\text{-}1}^v + (\mathbf{W}_{me} \mathbf{c}^m_t + \mathbf{W}_{ae} \mathbf{c}^a_t)  + \mathbf{b}_e)  \\
&\epsilon_{t,i} = \frac{\exp(e_{t,i})}{\sum_{j=1}^3 \exp(e_{t,j})} \text{ \quad  for } i=1\ldots 3 \\
\end{split}
\end{equation}
The memory $\mathbf{M}$ can be updated at each time step by
\begin{equation} 
\small
\mathbf{M}_t =  \epsilon_{t,1} \boldsymbol \alpha_{t}^m \mathbf{c}_t^m + \epsilon_{t,2} \boldsymbol \alpha_{t}^a \mathbf{c}_t^a + \epsilon_{t,3} \mathbf{M}_{t\text{-}1}
\end{equation} 
in which the write weights $\boldsymbol \alpha_t^{m/a}$ for memory slots determine how much attention should different slots pay to current inputs, while the modality weights $\boldsymbol \epsilon_t$ determine which of motion or appearance feature (or none of them if non-informational) from current inputs should the memory pay more attention to. Through this designed memory-write mechanism, we are able to integrate motion and appearance features to learn joint attention, and memorize different spatio-temporal patterns of this video in a synchronized and global context.


\textbf{Read operation}.
The next step is to perform an attentional read operation from the memory $\mathbf{M}$ to update memory hidden states.
We define the weights of reading from memory slots as $\small \boldsymbol \beta_t \text{=}\{\beta_{t,1},\dots,\beta_{t,S}\}$ given by
\begin{equation}\label{eq:softmax_b}
\small
\begin{split}
\mathbf{b}_t = &\mathbf{v}_b^\top \tanh( \mathbf{W}_{hb} \mathbf{h}_{t\text{-}1}^v + (\mathbf{W}_{mb} \mathbf{c}^m_t + \mathbf{W}_{ab} \mathbf{c}^a_t) + \mathbf{b}_b) \\
&\beta_{t,i} = \frac{\exp(b_{t,i})}{\sum_{j=1}^S \exp(b_{t,j})} \text{ \quad  for } i=1\ldots S  \\
\end{split}
\end{equation}
The content $\mathbf{r}_t$ read from memory is the weighted sum of each memory slot
$\mathbf{r}_t = \sum_{i=1}^S \beta_{t,i} \cdot \mathbf{m}_i$
in which both motion and appearance information has been integrated.

\textbf{Hidden states update}.
The final step is to update all three hidden states $\mathbf{h}^a$, $\mathbf{h}^m$ and $\mathbf{h}^{v}$
\begin{align} \label{eq:hidden}
\small
\mathbf{h}_{t}^{m/a} &= \sigma(\mathbf{W}_{hh}^{m/a} \mathbf{h}_{t\text{-}1}^{m/a} + \mathbf{W}_{oh}^{m/a} \mathbf{o}_t^{m/a} \qquad  \\
& \qquad  + \mathbf{W}_{rh}^{m/a} \mathbf{r}_t + \mathbf{b}_h^{m/a}) \nonumber \\
\mathbf{h}_{t}^{v} &= \sigma(\mathbf{W}_{hh}^{v} \mathbf{h}_{t\text{-}1}^{v} + \mathbf{W}_{rh}^{v} \mathbf{r}_t + \mathbf{b}_h^{v}) 
\end{align}
The global memory hidden state at all time steps $\mathbf{h}^v_{1:N_v}$ will be taken as our final video features.
In next section, we will discuss how to generate global question features. In
Section~\ref{sec:multimodal}, we will introduce how to interact video and question features for answer inference.

\begin{figure}
\vspace{-10pt}
\begin{center}
\includegraphics[width=.85\columnwidth, trim=6cm 4cm 5cm 4.2cm,clip]{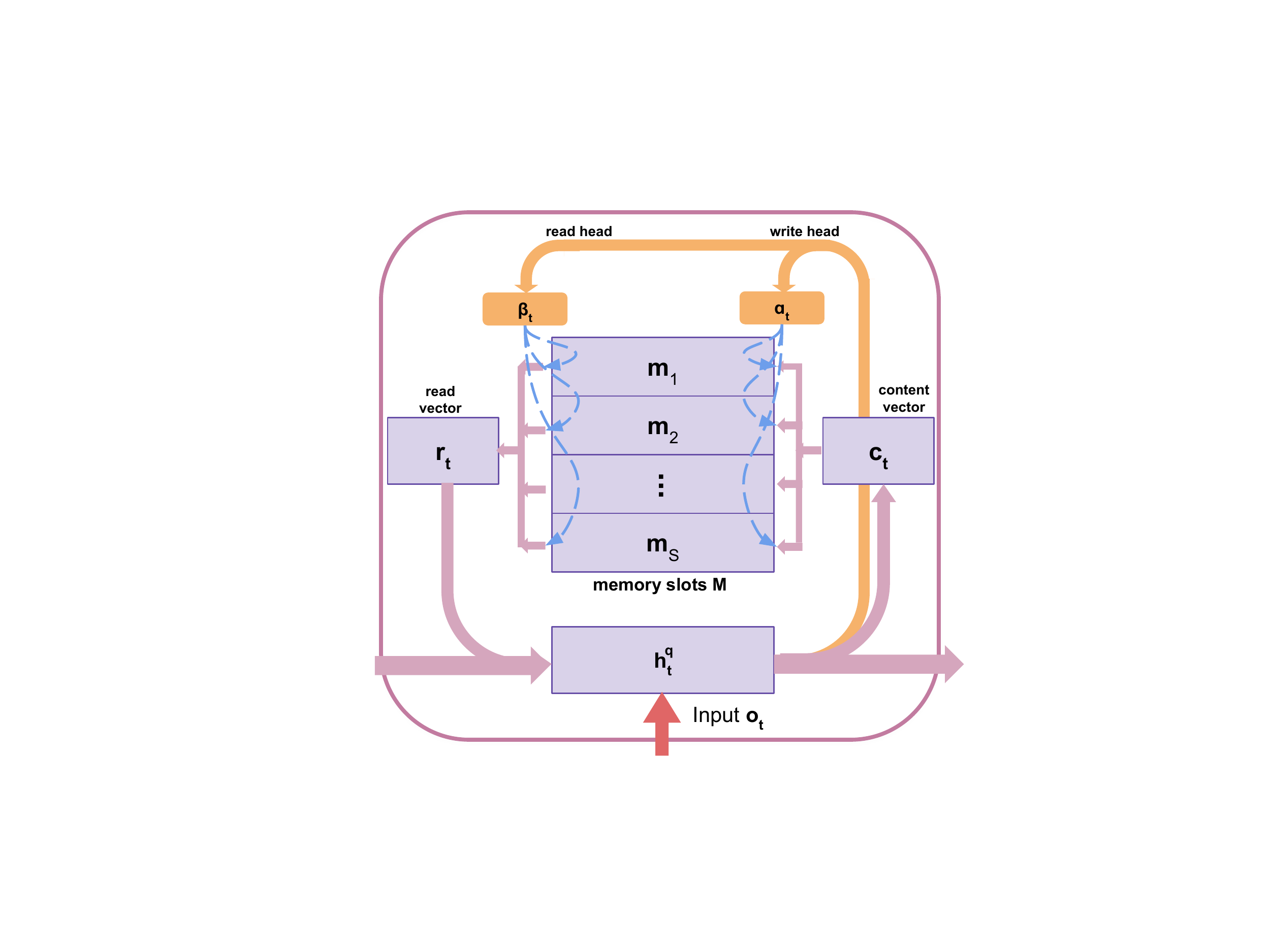}
\end{center}
\vspace{-10pt}
 \caption{Our re-designed question memory with memory slots $M$, read and write heads $\alpha,\beta$, and hidden states $\mathbf{h}^q$. }
\vspace{-10pt}
\label{fig:memory}
\end{figure}

\subsection{External question memory}

The existing deep learning based VideoQA methods often misunderstand the complex questions because they understand the questions based on local word information. For example, for question ``Who drives by a hitchhiking man who is smoking?", traditional methods are easily trapped by the local words and fail to generate the right attention to the queried person (the driver or the smoker). To address this challenging problem, we introduce the question memory to learn context-aware text knowledge. The question memory can store the sequential text information, learn relevance between words, and understand the question from the global point of view. 

We redesign the memory networks~\cite{graves2014neural,weston2015mm,sukhbaatar2015end,ma2018taxonomy} to persistently store previous inputs and enable interaction between current inputs and memory contents.
As shown in Fig.~\ref{fig:memory}, the memory module is composed of memory slots $\mathbf{M}=[\mathbf{m}_1,\mathbf{m}_2,\cdots,\mathbf{m}_S]$ and memory hidden state $\mathbf{h}^q$. 
Unlike the \htg memory discussed previously, one hidden state $\mathbf{h}^q$ is necessary for the question memory.
The inputs to the question memory are the encoded texts $\mathbf{o}^q$.

\textbf{Write operation}. We first define the content to write to the memory at $t$-th time step as $\mathbf{c}_t^q$ which is given by
\begin{equation} 
\small
\mathbf{c}_t^q = \sigma(\mathbf{W}_{oc} \mathbf{o}_t^q + \mathbf{W}_{hc} \mathbf{h}_{t\text{-}1}^q + \mathbf{b}_c)
\end{equation}
as a non-linear mapping from current input $\mathbf{o}_t^q$ and previous hidden state $\mathbf{h}^q_{t\text{-}1}$ to content vector $\mathbf{c}_t^q$.
Then we define the weights of writing to all memory slots
$\small \boldsymbol \alpha_t \text{=}\{\alpha_{t,1}...\alpha_{t,i}...\alpha_{t,S}\}$ such that
\begin{equation}  
\begin{split}
\small
\mathbf{a}_t &= \mathbf{v}_a^\top \tanh( \mathbf{W}_{ca} \mathbf{c}_t^q  + \mathbf{W}_{ha} \mathbf{h}_{t\text{-}1}^q + \mathbf{b}_a) \\
\alpha_{t,i} &= \frac{\exp(a_{t,i})}{\sum_{j=1}^S \exp(a_{t,j})} \text{ \quad  for } i=1\ldots S \\
\end{split}
\end{equation}
satisfying $\boldsymbol \alpha_t$ sum to $1$.
Then each memory slot $\mathbf{m}_i$ is updated by 
$\mathbf{m}_i =  \alpha_{t,i} \mathbf{c}_t + (1-\alpha_{t,i}) \mathbf{m}_i \text{   for } i=1\ldots S$.

\textbf{Read operation}. The next step is to perform attentional read operation from the memory slots $\mathbf{M}$. We define the normalized attention weights $\boldsymbol \beta_t \text{=}\{\beta_{t,1}...\beta_{t,i}...\beta_{t,S}\}$ of reading from memory slots such that
\begin{equation} \label{eq:softmax_b}
\begin{split}
\small
\mathbf{b}_t &= \mathbf{v}_b^\top \tanh( \mathbf{W}_{cb} \mathbf{c}_t^q + \mathbf{W}_{hb} \mathbf{h}_{t\text{-}1}^q +  \mathbf{b}_b) \\
\beta_{t,i} &= \frac{\exp(b_{t,i})}{\sum_{j=1}^S \exp(b_{t,j})} \text{ \quad  for } i=1\ldots S
\end{split}
\end{equation}
The content $\mathbf{r}_t$ read from memory is the weighted sum of each memory slot content $\small \mathbf{r}_t = \sum_{i=1}^S \beta_{t,i} \cdot \mathbf{m}_i$.

\textbf{Hidden state update}. The final step of $t$-th iteration is to update the hidden state $\mathbf{h}_t^q$ as 
\begin{equation} 
\small
\mathbf{h}_t^q = \sigma(\mathbf{W}_{oh} \mathbf{o}_t^q+ \mathbf{W}_{rh} \mathbf{r}_t + \mathbf{W}_{hh} \mathbf{h}_{t\text{-}1}^q + \mathbf{b}_h)
\end{equation}
We take the memory hidden state of all time steps $\mathbf{h}^q_{1:N_q}$ as the global context-aware question features which will be used for inference in Section~\ref{sec:multimodal}.

\subsection{Multimodal fusion and reasoning} \label{sec:multimodal}


In this section, we design a dedicated multimodal fusion and reasoning module for VideoQA, which can attend to multiple modalities such as visual and textual features, then make multi-step reasoning with refined attention for each modality. 
Our design is inspired by Hori~\etal~\cite{hori2017attention} which proposed to generate video captions by combining different types of features such as video and audio.

\begin{figure}
\vspace{-10pt}
\begin{center}
\includegraphics[width=0.85\columnwidth, trim=6cm 3.8cm 6cm 4.6cm,clip]{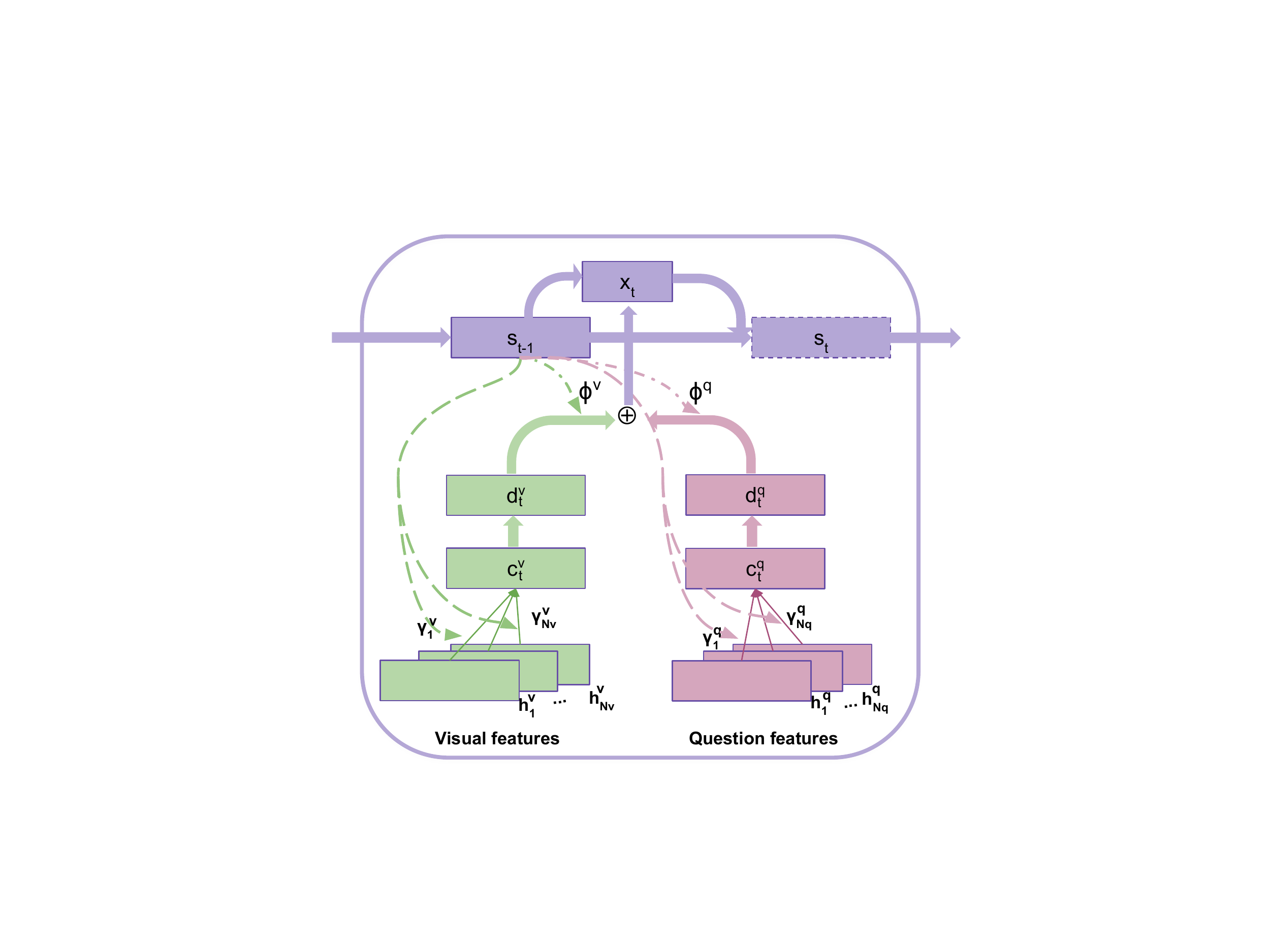}
\end{center}
\vspace{-10pt}
 \caption{Multimodal fusion layer. An LSTM controller with hidden state $\mathbf{s}_t$ attends to relevant visual and question features, and combines them to update current state.}
 \vspace{-10pt}
\label{fig:fusion}
\end{figure}

Fig.~\ref{fig:fusion} demonstrates our designed module. The hidden states of video memory $\mathbf{h}^v_{1:N_v}$ and question memory $\mathbf{h}^q_{1:N_q}$ are taken as the input features. The core part is an LSTM controller with its hidden state denoted as $\mathbf{s}$. During each iteration of reasoning, the controller attends to different parts of the video features and question features with temporal attention mechanism, and combines the attended features with learned modality weights $\boldsymbol \phi_t$, and finally updates its own hidden state $\mathbf{s}_t$. 

\textbf{Temporal attention.}
At $t$-th iteration of reasoning, we first generate two content vectors $\mathbf{c}_t^v$ and $\mathbf{c}_t^q$ by attending to different parts of visual features $\mathbf{h}_t^v$ and question features $\mathbf{h}_t^q$.
The temporal attention weights $\boldsymbol \gamma^v_{1:N_v}$ and $\boldsymbol \gamma^q_{1:N_q}$ are computed by
\begin{equation} \label{eq:temporal_attention}
\begin{split}
\small
\mathbf{g}^{v/q} &= \mathbf{v}_g^{^{v/q}\top} \tanh (\mathbf{W}^{v/q}_g \mathbf{s}_{t\text{-}1} + \mathbf{V}^{v/q}_g \mathbf{h}^{v/q} + \mathbf{b}^{v/q}_g) \\
& \gamma^{v/q}_{i} = \frac{\exp(g^{v/q}_i)}{\sum_{j=1}^{N_{v/q}} \exp(g^{v/q}_{j})} \text{ \quad  for } i=1\ldots N_{v/q} \\
\end{split}
\end{equation}
and shown by the dashed lines in Fig.~\ref{fig:fusion}.
Then the attended content vectors $\mathbf{c}_t^{v/q}$ and the transformed $\mathbf{d}_t^{v/q}$ are
\begin{equation} 
\small
\mathbf{c}^{v/q}_t = \sum_{i=1}^{N_{v/q}} \gamma^{v/q}_{i} \mathbf{h}^{v/q}_i, \,
\mathbf{d}^{v/q}_t = \text{ReLU}(\mathbf{W}_d^{v/q} \mathbf{c}^{v/q}_t + \mathbf{b}_d^{v/q})
\end{equation}

\textbf{Multimodal fusion.} The multimodal attention weights $\boldsymbol \phi_t=\{\phi_t^v,\phi_t^q \}$ are obtained by interacting the previous hidden state $\mathbf{s}_{t\text{-}1}$ with the transformed content vectors $\mathbf{d}^{v/q}_t$
\begin{equation}
\begin{split}
\small
\mathbf{p}_t^{v/q} &= \mathbf{v}_p^{\top} \tanh (\mathbf{W}^{v/q}_p \mathbf{s}_{t\text{-}1} + \mathbf{V}^{v/q}_p \mathbf{d}_t^{v/q} + \mathbf{b}_p^{v/q}) \\
&\phi_t^{v/q} = \frac{\exp(p_t^{v/q})}{\exp(p_t^{v}) + \exp(p_t^{q}) } \\
\end{split}
\end{equation}
The fused knowledge $\mathbf{x}_t$ is computed by the sum of $\mathbf{d}^{v/q}_t$ with multimodal attention weights $\phi^{v/q}$ such that 
\begin{equation} 
\small
\mathbf{x}_t = \phi_t^v \mathbf{d}^v_t + \phi_t^q \mathbf{d}^q_t
\end{equation}

\textbf{Multi-step reasoning.}
To complete $t$-th iteration of reasoning, the hidden state $\mathbf{s}_t$ of LSTM controller is updated by $\mathbf{s}_t = \text{LSTM}(\mathbf{x}_t,\mathbf{s}_{t\text{-}1})$.
This reasoning process is iterated for $L$ times and we set $L=3$. The optimal choice for $L$ is discussed in \S\ref{sec:abl_study}. The hidden state $\mathbf{s}_L$ at last iteration is the final representation of the distilled knowledge.
We also apply the standard temporal attention on encoded video features $\mathbf{o}^m$ and $\mathbf{o}^a$ as in ST-VQA~\cite{jang2017tgif}, and concatenate with $\mathbf{s}_L$ to form the final answer representation $\mathbf{s}_{A}$.



\subsection{Answer generation}
\label{sec:answer_generate}
We now discuss 
how to generate the correct answers from answer features $\mathbf{s}_{A}$.

\textbf{Multiple-choice} task is to choose one correct answer out of $K$ candidates. We concatenate the question with each candidate answer individually, and forward each QA pair to obtain the final answer feature $\{\mathbf{s}_A\}^{K}_{i=1}$, on top of which we use a linear layer to provide scores for all candidate answers
$\mathbf{s}=\{s^p,s^n_1,\cdots,s^n_{K-1}\}$ in which $s^p$ is the correct answer's score and the rest are $K-1$ incorrect ones. During training, we minimize the summed pairwise hinge loss~\cite{jang2017tgif} between the positive answer and each negative answer defined as
\begin{equation} 
\small
L_{mc}= \sum_{i=1}^{K-1} \max(0,m - (s^p - s^n_i))
\end{equation}
and train the entire network end-to-end.
The intuition of $L_{mc}$ is that the score of the true QA pair should be larger than any negative pair by a margin $m$. During testing, we choose the answer of highest score as the prediction. In Table~\ref{tab:data_stat}, we list the number of choices $K$ for each dataset.

\textbf{Open-ended} task is to choose one correct word as the answer from a pre-defined answer set of size $C$. We apply a linear layer and softmax function upon $\mathbf{s}_A$ to provide probabilities for all candidate answers such that $\mathbf{p} = \text{softmax} (\mathbf{W}^{\top}_p \mathbf{s}_L + \mathbf{b_p})$ in which $\mathbf{p} \in \mathbb{R}^C$.
The training error is measured by cross-entropy loss such that
\begin{equation} 
\small
L_{open} = - \sum_{c=1}^{C} \mathds{1}\{y=c\} \text{ } \log(p_c) 
\end{equation} 
in which $y$ is the ground truth label. By minimizing $L_{open}$ we can train the entire network end-to-end. 
In testing phase, the predicted answer is provided by $c^* = \argmax_c(\mathbf{p})$. 

\subsection{Implementation details}
We implemented our neural networks in PyTorch~\cite{paszke2017automatic} and updated network parameters by Adam solver~\cite{kingma2014adam} with batch size 32 and fixed learning rate $10^{-3}$. 
The video and question encoders are two-layer LSTMs with hidden size $512$. The dimension $D$ of the memory slot and hidden state is $256$.  We set the video and question memory sizes to 30 and 20 respectively, which are roughly equal to the maximum length of the videos and questions. We have released our code for boosting further research\footnote{https://github.com/fanchenyou/HME-VideoQA}.

%% file: experiments.tex
\section{Experiments and Discussions}
We evaluate our model on four benchmark VideoQA datasets and compare with the state-of-the-art techniques. 

\input{dataset}

\subsection{Result analysis}

\begin{table}[htb]
\centering
\setlength{\tabcolsep}{6pt}
    {\scriptsize{\textsf{
\begin{tabular}{@{}lccccccc@{}} \toprule  
 \multirow{2}{*}{\textbf{Method}}          & \multicolumn{4}{c}{\textbf{Question type}}   \\ \cmidrule{2-5} 
 & Count (loss)  & Action   & Trans.   & FrameQA                \\ \toprule
ST-VQA~\cite{jang2017tgif} & 4.28 & 0.608 & 0.671 & 0.493             \\  
Co-Mem~\cite{gao2018motion} & 4.10 & 0.682 & 0.743 & 0.515             \\ 
\midrule
Ours & \textbf{4.02} & \textbf{0.739} & \textbf{0.778} & \textbf{0.538} \\
                               \bottomrule
\end{tabular}
}}}
\vspace{3pt}
\caption{Experiment results on TGIF-QA dataset.}
\vspace{-3pt}
\label{tab:res_tgifqa}
\end{table}

\noindent\textbf{TGIF-QA result.}
Table~\ref{tab:res_tgifqa} summarizes the experiment results of all four tasks (Count,Action,Trans.,FrameQA) on TGIF-QA dataset. 
We compare with state-of-the-art methods ST-VQA~\cite{jang2017tgif} and Co-Mem~\cite{gao2018motion} and list the reported accuracy in the original paper.
For \textit{repetition counting} task (column 1), our method achieves the lowest average $L_2$ loss compared with ST-VQA and Co-Mem (4.02 v.s. 4.28 and 4.10). For \textit{Action} and \textit{Trans.} tasks (column 2,3), our method significantly outperforms the other two by increasing accuracy from prior best 0.682 and 0.743 to 0.739 and 0.778 respectively. For \textit{FrameQA} task (column 4), our method also achieves the best accuracy of 0.538 among all three methods, outperforming the Co-Mem by 4.7\%. 

\begin{table}[htb]
\centering
\setlength{\tabcolsep}{5pt}
    {\scriptsize{\textsf{
\begin{tabular}{@{}lcccccccc@{}} \toprule  
 \multirow{3}{*}{\textbf{Method}}          & \multicolumn{6}{c}{\textbf{Question type and \# instances}}   \\ \cmidrule{2-7} 
 & What  & Who   & How   & When  & Where & All                \\
 & 8419  & 4552   & 370   & 58  & 28 & 13427  \\ \toprule
ST-VQA~\cite{jang2017tgif} & 0.181 & \textbf{0.500} & \textbf{0.838} & 0.724 & 0.286 & 0.313    \\  
Co-Mem~\cite{gao2018motion} & 0.196 & 0.487 & 0.816 & \textbf{0.741} & 0.317 & 0.317 \\  
AMU~\cite{xu2017video} & 0.206 & 0.475 & 0.835 & 0.724 & \textbf{0.536} & 0.320   \\  
\cmidrule{1-7}
Ours & \textbf{0.224} &	\textbf{0.501} & 0.730 & 0.707 & 0.429 & \textbf{0.337} \\
                               \bottomrule
\end{tabular}
}}}
\vspace{3pt}
\caption{Experiment results on MSVD-QA dataset.}
\vspace{-3pt}
\label{tab:res_msvdqa}
\end{table}

\noindent\textbf{MSVD-QA result.}
Table~\ref{tab:res_msvdqa} summarizes the experiment results on MSVD-QA. 
It's worth mentioning that there is high class imbalance in both training and test sets,
as more than 95\% questions are \textit{what} and \textit{who} while less than 5\% are \textit{how}, \textit{when} and \textit{where}. We list the numbers of their test instances in the table for reference. 
We compare our model with the ST-VQA~\cite{jang2017tgif}, Co-Mem~\cite{gao2018motion} and current state-of-the-art AMU~\cite{xu2017video} on MSVD-QA. We show the reported accuracy of AMU in~\cite{xu2017video}, while we accommodate
the source code of ST-VQA and implement Co-Mem from scratch to obtain their numbers.
Our method outperforms all the others on both \textit{what} and \textit{who} tasks, and achieves best overall accuracy of 0.337 which is 5.3\% better than prior best (0.320). Even though our method slightly underperforms on the How, When and Where questions, the difference are minimal (40,2 and 3) regarding the absolute number of instances due to class imbalance.

\begin{table}[htb]
\centering
\setlength{\tabcolsep}{4pt}
    {\scriptsize{\textsf{
\begin{tabular}{@{}lcccccccc@{}} \toprule  
 \multirow{2}{*}{\textbf{Method}}          & \multicolumn{6}{c}{\textbf{Question type}}   \\ \cmidrule{2-7} 
 & What  & Who   & How   & When  & Where & All                \\ \toprule
ST-VQA~\cite{jang2017tgif} & 0.245 & 0.412 & 0.780 & \textbf{0.765} & 0.349 & 0.309    \\  
Co-Mem~\cite{gao2018motion} & 0.239 & 0.425 & 0.741 & 0.690 & \textbf{0.429} & 0.320  \\  
AMU~\cite{xu2017video} &  0.262	&0.430	&0.802	&0.725	&0.300	&0.325           \\  
\cmidrule{1-7}
Ours &\textbf{0.265} &\textbf{0.436} &\textbf{0.824}	&0.760	&0.286	&\textbf{0.330} \\
                               \bottomrule
\end{tabular}
}}}
\vspace{3pt}
\caption{Experiment results on MSRVTT-QA dataset.}
\vspace{-3pt}
\label{tab:res_msrvtt}
\end{table}

\noindent\textbf{MSRVTT-QA result.}
In Table~\ref{tab:res_msrvtt}, we compare our model with the ST-VQA~\cite{jang2017tgif}, Co-Mem~\cite{gao2018motion} and AMU~\cite{xu2017video} on MSRVTT-QA.
Similar to the trend on MSVD-QA, 
our method outperforms the other models on three major question types (\textit{what}, \textit{who}, \textit{how}), and achieves the best overall accuracy of 0.330.

\begin{table}[htb]
\centering
\setlength{\tabcolsep}{4pt}
    {\scriptsize{\textsf{
\begin{tabular}{@{}lcccccccc@{}} \toprule  
\multirow{3}{*}{\textbf{Task}}  & \multirow{2}{*}{\textbf{Method}}          & \multicolumn{5}{c}{\textbf{Question type and \# instances}}   \\ \cmidrule{3-7} 
\multicolumn{2}{c}{}                                   & What  & Who   & Other   & All  & Avg. Per-class                \\
 & & 2489 & 2004 & 97 & 4590\\ \toprule
\multirow{2}{*}{Multi-choice}    & r-ANL~\cite{ye2017video}   &0.633	&0.364	&0.845	&0.520	&0.614           \\  
\cmidrule{2-7} 
& Ours &\textbf{0.831}	&\textbf{0.778}	&\textbf{0.866}	&\textbf{0.808}	&\textbf{0.825}\\
\midrule
\multirow{2}{*}{Open-ended}  &
r-ANL~\cite{ye2017video}  &0.216	&\textbf{0.294}	&\textbf{0.804}	&0.262	&0.438             \\  
\cmidrule{2-7}
& Ours &\textbf{0.292}	&0.287	&0.773	&\textbf{0.301}	&\textbf{0.451}\\
                               \bottomrule
\end{tabular}
}}}
\vspace{-3pt}
\caption{Experiment results on YouTube2Text-QA dataset.}
\vspace{-3pt}
\label{tab:res_zhqa}
\end{table}

\noindent\textbf{YouTube2Text-QA result.}
In Table~\ref{tab:res_zhqa}, we compare our methods with the state-of-the-art r-ANL~\cite{ye2017video} on YouTube2Text-QA dataset. It's worth mentioning that r-ANL utilized frame-level attributes as additional supervision to augment learning while our method does not.
For multiple-choice questions, our method significantly outperforms r-ANL on all three types of questions (\textit{What, Who, Other}) and achieves a better overall accuracy (0.808 v.s. 0.520). 
For open-ended questions, our method outperforms r-ANL on \textit{what} queries and slightly underperforms on the other two types. Still, our method achieves a better overall accuracy (0.301 v.s. 0.262).
We also report the per-class accuracy 
to make direct comparison with~\cite{ye2017video}, and our method is better than r-ANL in this evaluation method.

\subsection{Attention visualization and analysis}
\begin{figure}[t]
\begin{center}
\includegraphics[width=1.0\columnwidth, trim=4.8cm 2.3cm 4.0cm 1.4cm,clip]{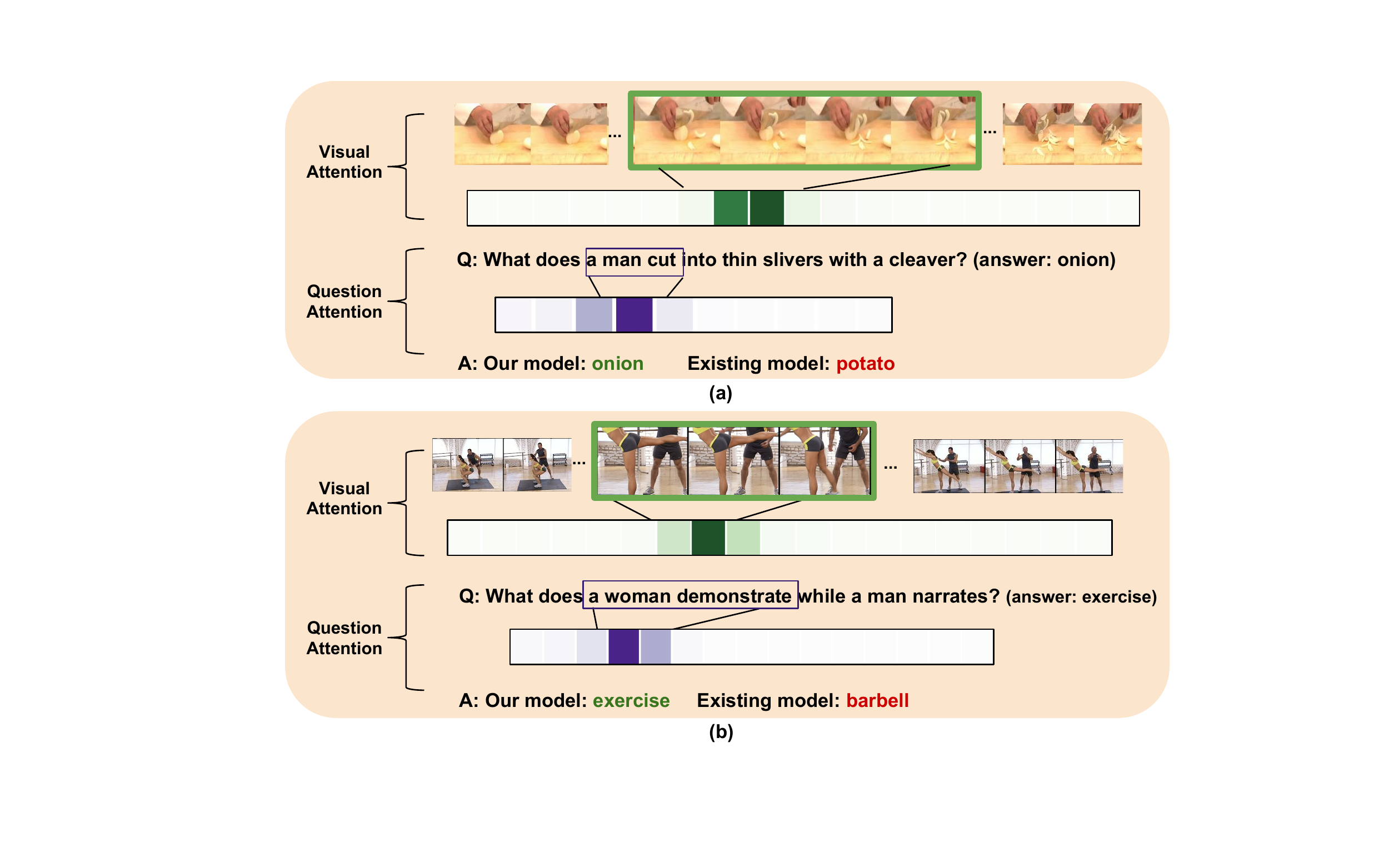}
\end{center}
\vspace{-10pt}
 \caption{Visualization of multimodal attentions learned by our model on two QA exemplars. Highly attended frames and words are highlighted.}
\label{fig:samples}
\end{figure}

In Figs.~\ref{fig:demo} and~\ref{fig:samples}, we demonstrate three QA examples with highlighted key frames and words which are recognized by our designed attention mechanism. For visualization purpose, we extract the visual and textual attention weights from our model (Eq.~\ref{eq:temporal_attention}) and plot them with bar charts. Darker color stands for larger weights, showing that the corresponding frame or word is relatively important.

Fig.~\ref{fig:demo} shows the effectiveness of understanding complex question with our proposed question memory. This question intends to query the female driver though it uses another relative clause to describe the man. Our model focuses on the correct frames in which the female driver is driving in the car and also focuses on the words which describe the woman but not the man. In contrast, ST-VQA~\cite{jang2017tgif} fails to identify the queried person as its simple temporal attention is not able to gather semantic information in the context of a long sentence.

In Fig.~\ref{fig:samples}(a), we provide an example showing that our video memory is learning the most salient frames for the given question while ignoring others. In the first half of the video, it's difficult to know whether the vegetable is onion or potato, due to the lighting condition and camera view. However, our model smartly pays attention to frames in which the onion is cut into pieces by combining both question words ``a man cut" and the motion features, and thus determines the correct object type by onion pieces (but not potato slices) from appearance hint. 

Fig.~\ref{fig:samples}(b) shows a typical example illustrating that jointly learning motion and appearance features as our \htg memory design is superior to attending to them separately such as Co-Mem~\cite{gao2018motion}. In this video, a woman is doing yoga in a gym, and there is a barbell rack at the background. Our method successfully associated the woman with the action of exercising, while Co-Mem~\cite{gao2018motion} incorrectly pays attention to the barbell and fails to utilize motion information as they separately learn motion and appearance attentions.

\subsection{Ablation study}
\label{sec:abl_study}
We perform two ablation studies to investigate the effectiveness of each component of our model. We first study how many iterations of reasoning is sufficient in the designed multimodal fusion layer. After that, we make a comparison of variants of our model to evaluate the contribution of each component.

\noindent\textbf{Reasoning iterations.}
To understand how many iterations of reasoning are sufficient for our VideoQA tasks, we test different numbers and report their accuracy.
The validation accuracy on MSVD-QA dataset increases from 0.298 to 0.306 when the number of reasoning iteration $L$ increases from 1 to 3, and seems to saturate at $L=5$ (0.307), while drops to 0.304 at $L=7$. To balance performance and speed, we choose $L=3$ for our experiments throughout the paper.



\begin{table}[htb]
\centering
\small
\begin{tabular}{|c|c|c|c|c|c|c|}
\hline
Dataset & EF &LF & E-M & V-M & Q-M & V+Q \\ \hline
MSVD   &0.313 & 0.315 & 0.318  & 0.320  & 0.315   &\textbf{0.337}     \\ \hline
MSRVTT &0.309 & 0.312 & 0.319  & 0.325  & 0.321 & \textbf{0.330}    \\ \hline
\end{tabular}
 \caption{Ablation study of different architectures.}
 \label{tab:ablation}
\end{table}

\noindent\textbf{Different architectures.}
To understand the effectiveness of our designed memory module, we compare several variants of our models and evaluate on MSVD-QA and MSRVTT-QA, as shown in Table~\ref{tab:ablation}. 
\textit{Early Fusion (EF)} is indeed ST-VQA~\cite{jang2017tgif} which concatenates raw video appearance and motion features at an early stage, before feeding into the LSTM encoder.
\textit{Late Fusion (LF)} model uses two separate LSTM encoders to encode video appearance and motion features and then fuses them by concatenation. 
\textit{Episodic Memory (E-M)}~\cite{xiong2016dynamic} is a simplified memory network embodiment and we use it as the visual memory to compare against our design.
\textit{Visual Memory (V-M)} model uses our designed \htg visual memory ($M^v$ in Fig.~\ref{fig:pipeline}) to fuse appearance and motion features and generate global context-aware video features. 
\textit{Question Memory (Q-M)} model uses our redesigned question memory only ($M^q$ in Fig.~\ref{fig:pipeline}) to better capture complex question semantics.
Finally, \textit{Visual and Question Memory (V+Q M)} is our full model which has both visual and question memory. 

In Table~\ref{tab:ablation}, we observe consistent trend that using memory networks (e.g., E-M,V-M,V+Q) to align and integrate multimodal visual features is generally better than simply concatenating them (e.g., EF,LF). In addition,  our designed visual memory (V-M) has shown its strengths over episodic memory (E-M) and other memory types (Table~\ref{tab:res_msvdqa}-\ref{tab:res_zhqa}). Furthermore, using both visual memory and question memory (V+Q) increases the performance by 2-7\%.

%% file: dataset.tex
\begin{table*}[ht]
\vspace{-8pt}
\centering
\setlength{\tabcolsep}{4pt}
    {\scriptsize{\textsf{
\begin{tabular}{@{}lccccccccc@{}} \toprule  
\multirow{2}{*}{\textbf{Dataset}}  & \multirow{2}{*}{\textbf{Feature}}  & \multirow{2}{*}{\textbf{Vocab size}}  &  \multirow{2}{*}{\textbf{Video len}} &  \multirow{2}{*}{\textbf{Video num}} & \multicolumn{3}{c}{\textbf{Question num}}  & \textbf{Ans size} & \textbf{MC num} \\ \cmidrule{6-8} 
\multicolumn{5}{c}{}    & Train  & Val   & Test        \\ \toprule
TGIF-QA~\cite{jang2017tgif} & ResNet+C3D & 8,000 & 35 & 71,741 & 125,473 & 13,941 & 25,751 & 1746 & 5\\
MSVD-QA~\cite{xu2017video} & VGG+C3D & 4,000 & 20 & 1,970 & 30,933 & 6,415 & 13,157  & 1000 & NA\\
MSRVTT-QA~\cite{xu2017video} & VGG+C3D & 8,000 & 20 & 10,000 & 158,581 & 12,278 & 72,821 & 1000 & NA\\
Youtube2Text-QA~\cite{ye2017video} & ResNet+C3D & 6,500 & 40 & 1,970 & 88,350 & 6,489 & 4,590 & 1000 & 4\\
\bottomrule
\end{tabular}
}}}
\vspace{3pt}
\caption{Dataset statistics of four VideoQA benchmark datasets. The columns from left to right indicate dataset name, feature types, vocabulary size, sampled video length, number of videos, size of QA splits, answer set size (Ans size) for open-ended questions, and number of options for multiple-choice questions (MC num).}
\vspace{-3pt}
\label{tab:data_stat}
\end{table*}

\subsection{Dataset descriptions}
\label{sec:dataset}

In Table~\ref{tab:data_stat}, we show the statistics of the four VideoQA benchmark datasets and the experimental settings from their original paper including feature types, vocabulary size, sampled video length, number of videos, size of QA splits, answer set size for open-ended questions, and number of options for multiple-choice questions. 

\textbf{TGIF-QA}~\cite{jang2017tgif} contains 165K QA pairs associated with 72K GIF images based on the TGIF dataset~\cite{li2016tgif}. TGIF-QA includes four types of questions: 1) counting the number of occurrences of a given action; 2) recognizing a repeated action given its count; 3) identifying the action happened before or after a given action, and 4) answering image-based questions.
\textbf{MSVD-QA} and \textbf{MSRVTT-QA} were proposed by
Xu~\etal~\cite{xu2017video} based on MSVD~\cite{chen2011collecting} and MSVTT~\cite{Xu:CVPR16} video sets respectively. Five different question types exist in both datasets, including what, who, how, when and where. The questions are open-ended with pre-defined answer sets of size 1000.
\textbf{YouTube2Text-QA} ~\cite{ye2017video} collected three types of questions (what, who and other) from the YouTube2Text~\cite{guadarrama2013youtube2text} video description corpus. The video source is also MSVD~\cite{chen2011collecting}. Both open-ended and multiple-choice tasks exist. 



%% file: conclusion.tex
\section{Conclusion}
In this paper, we proposed a novel end-to-end deep learning framework for VideoQA, with designing new external memory modules to better capture global contexts in video frames, complex semantics in questions, and their interactions. 
A new multimodal fusion layer was designed to fuse visual and textual modalities and perform multi-step reasoning with gradually refined attention. In empirical studies, we 
visualized the attentions generated by our model to verify its capability of understanding complex questions and attending to salient visual hints. Experimental results on four benchmark VideoQA datasets show that our new approach consistently outperforms state-of-the-art methods.